\pdfoutput=1
\documentclass[11pt]{article}

\usepackage[final]{acl}
\usepackage{times}
\usepackage{latexsym}
\usepackage[T1]{fontenc}
\usepackage[utf8]{inputenc}
\usepackage{microtype}
\usepackage{inconsolata}
\usepackage{graphicx}
\usepackage{subfigure}
\usepackage{amsmath}
\usepackage{CJKutf8}
\usepackage{multirow}
\usepackage{pinyin}
\usepackage{amssymb}
\usepackage{xcolor}
\usepackage{amsmath}
\usepackage{amsfonts}
\usepackage{algorithm}
\usepackage{algorithmic}
\newcommand{\zh}[1]{\begin{CJK*}{UTF8}{gbsn}#1\end{CJK*}}
\newcommand{\method}{OBSD}
\title{Deciphering Oracle Bone Language with Diffusion Models}

\author{Haisu Guan\textsuperscript{\textmd{1}}, Huanxin Yang\textsuperscript{\textmd{1}}, Xinyu Wang\textsuperscript{\textmd{2}}, Shengwei Han\textsuperscript{\textmd{3}}, Yongge Liu\textsuperscript{\textmd{3}}, \\
{\bf Lianwen Jin\textsuperscript{\textmd{4}}, Xiang Bai\textsuperscript{\textmd{1}}, Yuliang Liu\textsuperscript{\textmd{1,*}}} \\
\textsuperscript{1}Huazhong University of Science and Technology\quad \textsuperscript{2}The University of Adelaide \\
\textsuperscript{3}Anyang Normal University\quad \textsuperscript{4}South China University of Technology \\
\texttt{\textsuperscript{1}\{haisuguan, ylliu\}@hust.edu.cn} \\
\texttt{\textsuperscript{*}Corresponding author}}

\begin{document}
\maketitle
\begin{abstract}
Originating from China's Shang Dynasty approximately 3,000 years ago, the Oracle Bone Script (OBS) is a cornerstone in the annals of linguistic history, predating many established writing systems. Despite the discovery of thousands of inscriptions, a vast expanse of OBS remains undeciphered, casting a veil of mystery over this ancient language. The emergence of modern AI technologies presents a novel frontier for OBS decipherment, challenging traditional NLP methods that rely heavily on large textual corpora, a luxury not afforded by historical languages. This paper introduces a novel approach by adopting image generation techniques, specifically through the development of Oracle Bone Script Decipher (\method{}). Utilizing a conditional diffusion-based strategy, \method{} generates vital clues for decipherment, charting a new course for AI-assisted analysis of ancient languages. To validate its efficacy, extensive experiments were conducted on an oracle bone script dataset, with quantitative results demonstrating the effectiveness of \method{}. \textit{Code and decipherment results will be made available} at \href{https://github.com/guanhaisu/OBSD}{https://github.com/guanhaisu/OBSD}.
\end{abstract}

\section{Introduction}

Oracle Bone Script (OBS) represents an ancient language inscribed on turtle shells and animal bones, extensively utilized during China's Shang Dynasty, a feudal dynasty dating back 3,000 years. The script not only chronicled the human geography and daily activities of that period but also encapsulates invaluable historical significance, offering a unique window into the linguistic and cultural practices of early Chinese civilization. However, despite the discovery of tens of thousands of fragments of oracle bones, a significant portion of the characters remain undeciphered~\cite{wang2024dataset}, leaving the rest shrouded in mystery. To date, more than 4,500 Oracle Bone Script (OBS) characters have been discovered, but only about 1,600 of these have been deciphered and linked to their modern Chinese counterparts. In modern Chinese, Unicode includes more than 90,000 Chinese characters, though only approximately 3,500 characters are commonly used in contemporary Chinese society. This challenge of understanding the remaining undeciphered OBS characters and linking them to modern Chinese has attracted significant research interest, with attempts being made to leverage modern AI technologies for the understanding of such an ancient language~\cite{zhang2022data, jiang2023oraclepoints, wang2024dataset, guan2024open}.

However, the majority of existing methodologies primarily focus on the recognition and understanding of already deciphered OBS~\cite{guo2015building, meng2018recognition, zhang2019oracle, hu2023coding}, with the utilization of AI to assist in the decipherment of unknown inscriptions remaining an underexplored area. This is partly because, unlike modern languages that can be digitized and stored as text due to established encoding systems, OBS lacks a standard input method or encoding scheme, resulting in its preservation predominantly in the form of images rather than digital text usually used in NLP methods. Additionally, since OBS was inscribed on turtle shells and animal bones, many of which have been damaged or fragmented upon discovery, there is essentially no complete corpus available. This absence of a comprehensive corpus severely limits the applicability of language models that require extensive datasets for training, such as BERT~\cite{devlin2018bert}, RoBERTa~\cite{liu2019roberta}, and GPT~\cite{brown2020language}.

To address the challenges inherent in the decipherment of OBS using conventional NLP methodologies, this paper introduces a novel approach by employing image-based generative techniques for auxiliary decipherment of OBS. Specifically, we train a conditional diffusion model that utilizes unseen categories of OBS as a conditional input to generate corresponding images of its modern counterpart. This direct provision of modern representations or potential decipherment clues leverages the model's learned evolution from ancient scripts to contemporary fonts, circumventing the corpus construction and other challenges that traditional NLP methods face with ancient languages. Notably, while our experiments focus on OBS, this training paradigm holds the potential for extension to other ancient languages, such as Cuneiform and Hieroglyphics. In summary, this paper makes three key contributions:

\begin{itemize}
    \item We introduce a novel approach to the task of ancient script decipherment by utilizing image generation techniques, offering a novel solution to challenges that conventional NLP methods struggle to address.
    \item We propose Oracle Bone Script Decipher (OBSD), a conditional diffusion model optimized for OBS decipherment. Our Localized Structural Sampling technique enhances the model's ability to discern and interpret the intricate patterns of characters.
    \item OBSD demonstrates its effectiveness in decipherment through comprehensive ablation studies and benchmark comparisons. It offers a pioneering approach for AI-assisted ancient language decipherment, potentially laying a foundation for future research.
\end{itemize}

\section{Related Works}

Applying machine learning to the study of ancient languages represents a notable shift in linguistics and epigraphy. This area, distinct from the NLP tasks typically associated with modern languages, involves digitization, linguistic analysis, textual criticism, translation, and decipherment~\cite{jin2023morphological, nuhn2012deciphering, ravi2011deciphering}. For a comprehensive overview of this field, we direct interested readers to the survey by Sommerschield et al.~\cite{sommerschield2023machine, li2020hwobc, huang2019obc306, yang2020oracle, guo2015building}. Due to space constraints, our review is limited to literature most pertinent to oracle bone language decipherment.

The oracle bone language is considered a form of hieroglyphic that uses pictorial symbols to represent specific meanings. It originated around 1500 BC and has evolved over thousands of years into modern Chinese characters. The evolution timeline can be summarized into seven periods as follows: Oracle Bone Script (1500 BC), Bronze Inscriptions (1300 BC - 221 BC), Seal Script (1100 BC - 221 BC), Spring \& Autumn Characters (770 BC - 476 BC), Warring States Characters (475 BC - 221 BC), Clerical Script (221 BC - 220 AD) and Regular Script (around 3rd century AD). The continuous evolutionary path makes OBS a unique presence among ancient scripts. Many of its character forms have been preserved in modern standard Chinese characters. While these are significant overlaps in the forms and meanings of characters between adjacent periods, greater differences can be found between more distant periods. Some characters disappeared and later reappeared across different periods, highlighting the dynamic nature of this ancient writing system.

While the majority of work related to OBS has focused on employing CV or NLP techniques to recognize~\cite{zhang2021ai, fu2022improvement, wang2022unsupervised} or understand~\cite{han2020isobs, qi2023vector, hu2023coding} already deciphered characters, the use of AI to assist in deciphering characters with unknown meanings remains a largely unexplored and challenging task. Among these, the case-based reasoning strategy developed by Zhang et al.~\cite{zhang2021deciphering} stands out in its method of drawing parallels to already interpreted characters to decipher OBS. While effective to a degree, this approach is inherently constrained by its dependence on the corpus of previously deciphered characters, potentially stymieing the discovery of novel meanings. On another front, Chang et al.'s cascade generative adversarial networks framework~\cite{chang2022sundial} presents an innovative attempt at deciphering, yet it faces challenges due to evolutionary gaps in OBS and the completeness of training data. These challenges arise because when a character disappears for a specific period, the evolutionary path relied upon by such methods no longer remains intact, significantly impacting the success rate of deciphering and restricting their effectiveness to small datasets with clear evolutionary paths.

\section{Method}

\subsection{Preliminary}
\label{sec:preliminary}

\begin{figure*}[ht]
\centering
\includegraphics[width=0.8\linewidth]{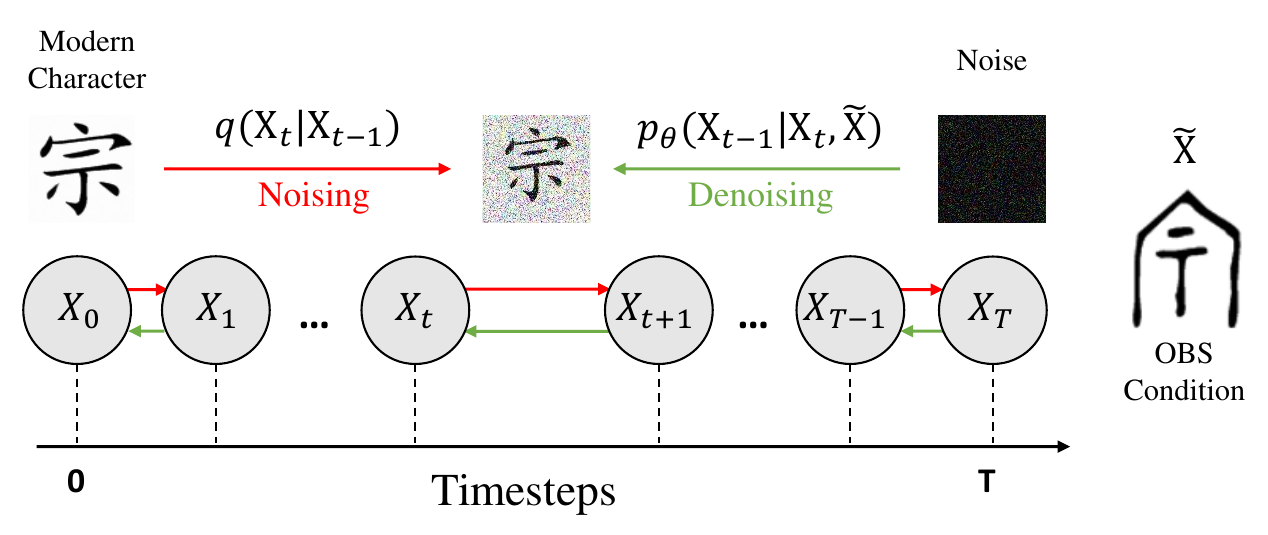}
\caption{Conditional diffusion model for OBS decipherment.}
\label{fig:OBC_Diffusion}
\end{figure*}

In this study, we focus on the task of OBS decipherment, aiming to predict the corresponding modern Chinese character forms for the oracle bone language. This endeavor not only seeks to match known characters but also to uncover new forms that could elucidate the meanings of these ancient scripts. Formally, the training set denoted as $S = \{(s_{i}, c_{i}) \mid s_{i} \text{ is an OBS instance and } c_{i} \in C\}$, pairs OBS instances with their modern Chinese counterparts from a set of known Categories $C$. The model is designed to extend beyond the training set $S$, identifying modern equivalents for OBS instances $s'$, and proposing new character forms where existing matches are absent.

To achieve this, our approach utilizes a diffusion-based~\cite{ho2020denoising} model, for transforming OBS character images $\tilde{X}$ into their modern Chinese equivalents, as illustrated in Figure~\ref{fig:OBC_Diffusion}. The model operates in two phases: the forward process, the forward phase introduces noise to the modern Chinese character images $X_{0}$, transitioning them towards a state resembling pure noise via a controlled Markov chain process, ultimately conforming to a Gaussian distribution $\mathcal{N}(0, I)$. This is mathematically articulated as follows:
\begin{equation}
    q\left(X_{1:T}\mid X_{0}\right)=\prod_{t=1}^{T}q\left(X_{t}\mid X_{t-1}\right)
\end{equation}

\noindent where $T$ denotes the total number of steps. For each step $t$, noise is added according to the following equation:
\begin{equation}
\resizebox{0.85\linewidth}{!}{$
    q\left(X_t\mid X_{t-1}\right)= \mathcal{N}\left(X_t\mid\sqrt{\alpha_t}X_{t-1},\left(1-\alpha_t\right)I\right)
$}
\end{equation}

\noindent where $\alpha_{t}$ is a hyperparameter controlling the noise intensity, and $I$ represents the identity matrix. The transition from $X_{0}$ to a noisy state $X_t$ over $t$ step is captured by the equation:
\begin{equation}
    X_t=\sqrt{\gamma_t}X_0+\sqrt{1-\gamma_t}\epsilon,\quad\epsilon\sim\mathcal{N}(0,I)
\end{equation}

\noindent with $\gamma_{t}$ being the cumulative product of $\alpha$ values up to $t$. 

The denoising phase employs a U-Net architecture~\cite{ronneberger2015u} for the model $f_{\theta}$, trained to predict the noise $\epsilon$ and restore the image. The training objective minimizes the loss function:
\begin{equation}
    \mathcal{L}=\mathbb{E}_{\epsilon, \gamma} \left\|\epsilon-f_\theta\left(\tilde{X},X_t,\gamma\right)\right\|^2
\label{eq:loss}
\end{equation}

\noindent which measures the discrepancy between the actual noise $\epsilon$ and its estimation by the $f_{\theta}$. In the inference stage $p_{\theta}(X_t\mid X_t,\tilde{X})$, we reverse the noise addition process, starting from the noisiest state $X_T$ and iteratively denoising down to $t=1$.
\begin{equation}
\resizebox{0.85\linewidth}{!}{$
    X_{t-1} = \frac1{\sqrt{\alpha_t}}\left(X_t - \frac{1-\alpha_t}{\sqrt{1-\gamma_t}}f_\theta\left(\tilde{X}, X_t,\gamma_t\right)\right) + \sqrt{1 - \alpha_t}\epsilon_t
    \label{eq:sample}
$}
\end{equation}

\noindent where $\epsilon_t\sim\mathcal{N}(0,I)$ introduces randomness to enhance the diversity of model generated results. The outcome is the denoised image $\hat{X}_{0}$, representing the deciphered results.

\begin{figure*}[t!]
\centering
\includegraphics[width=0.8\linewidth]{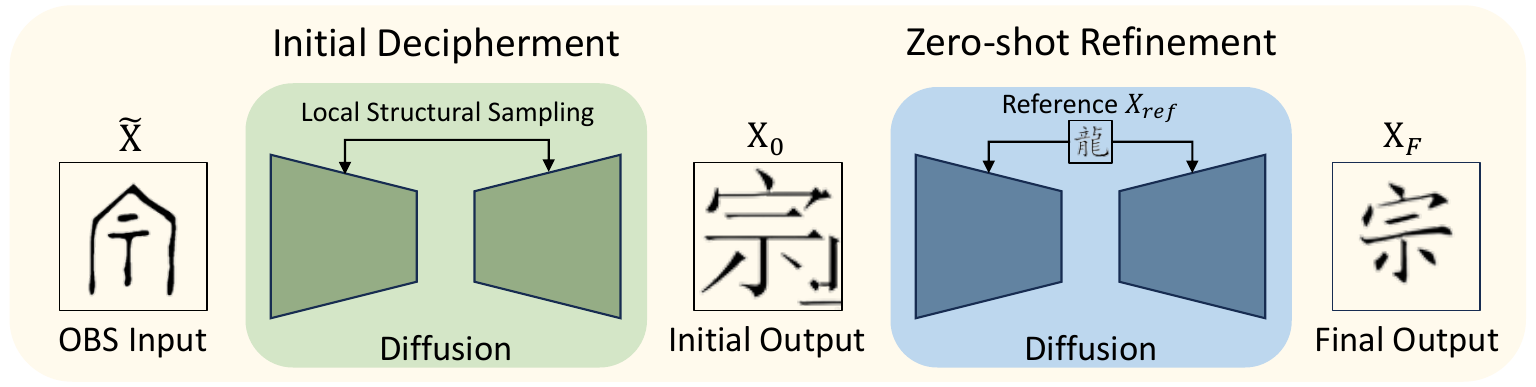}
\caption{Overview pipeline of the proposed \method{}. The input OBS $\tilde{X}$ undergoes a diffusion model to generate initial decipherment result $X_0$, which is then refined with a style-specific reference to produce the final output $X_F$.}
\label{fig:OBCdiffuser}
\end{figure*}

Building on this, our \method{} model integrates an Initial Decipherment phase with a Zero-shot Refinement stage to improve the decipherment accuracy. As shown in Figure~\ref{fig:OBCdiffuser}, initially, an OBS image $\tilde{X}$ undergoes conditional diffusion to approximate an initial decipherment $X_{0}$, which is then refined using a zero-shot learning approach, leveraging a reference style image $X_{ref}$ to correct and enhance the structure. with a distinct style to enhance $X_{0}$, learning from the structure of modern Chinese characters. The final result $X_{F}$ emerges as a refined representation of the intended modern Chinese character, benefiting from the refinement process's structural insights.

\subsection{Initial Decipherment}

\begin{figure}[t]
    \centering
    \includegraphics[width=1\linewidth]{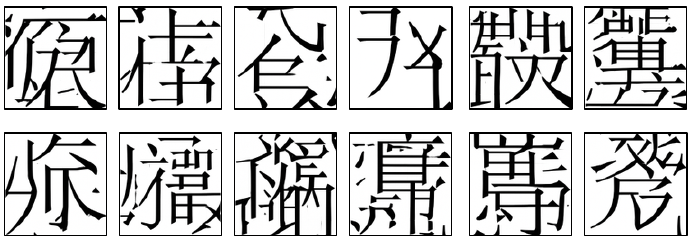}
    \caption{Directly training a conditional diffusion model results in \textbf{failure} decipherment.}
    \label{fig:failure}
\end{figure}

\begin{figure}[t]
    \centering
    \subfigure[Global Structure]{\includegraphics[width=0.45\linewidth]{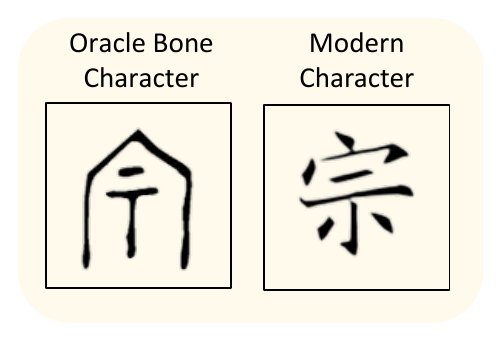}\label{fig:global}}
    \subfigure[Local Structure]{\includegraphics[width=0.45\linewidth]{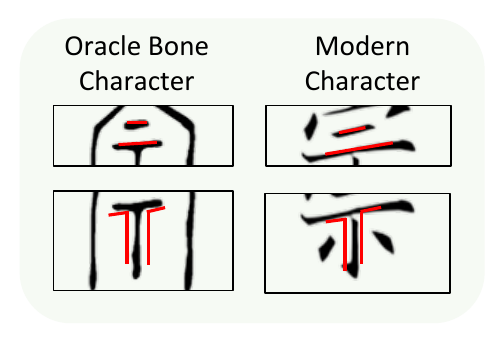}\label{fig:local}}
    \caption{Comparative analysis of the Chinese character \zh{宗} (\zong1): (a) Depicts the evolution of global structure from OBS to its modern form. (b) Highlights the retention of specific local structures amidst the evolution.}
    \label{fig:global-local}
    \vspace{-0.3cm}
\end{figure}

After revisiting the fundamentals in Section~\ref{sec:preliminary}, a preliminary and somewhat naive idea was to directly utilize OBS images as the condition $\tilde{X}$ and modern Chinese characters as the target images $X_{0}$ to train a conditional diffusion model for decipherment. However, as shown in Figure~\ref{fig:failure}, we observed that directly training such a model did not result in the accurate generation of the corresponding photos of modern Chinese characters. Instead, the model produces images comprised of a multitude of random stroke fragments, resembling gibberish. We speculate that this discrepancy arises because diffusion models are primarily designed for generating natural images, where the input conditions, such as edges and sketches, provide structural information to guide the generation of target images. However, in the context of deciphering OBS, the structural disparity between the input OBS images and the expected modern Chinese character outcomes is significant (see Figure~\ref{fig:global}), rendering the standard conditional diffusion model ineffective for accurate reconstruction of the target modern characters. To address this challenge, we introduce the concept of Localized Structural Sampling (LSS) as a means to aid the diffusion model in learning how to map local radical structures of OBS to the corresponding modern Chinese character space (see Figure~\ref{fig:local} red marks), thereby enhancing the model's capability to bridge the structural gap between ancient inscriptions and contemporary linguistic forms.

\begin{figure*}[t!]
\centering
\includegraphics[width=0.8\linewidth]{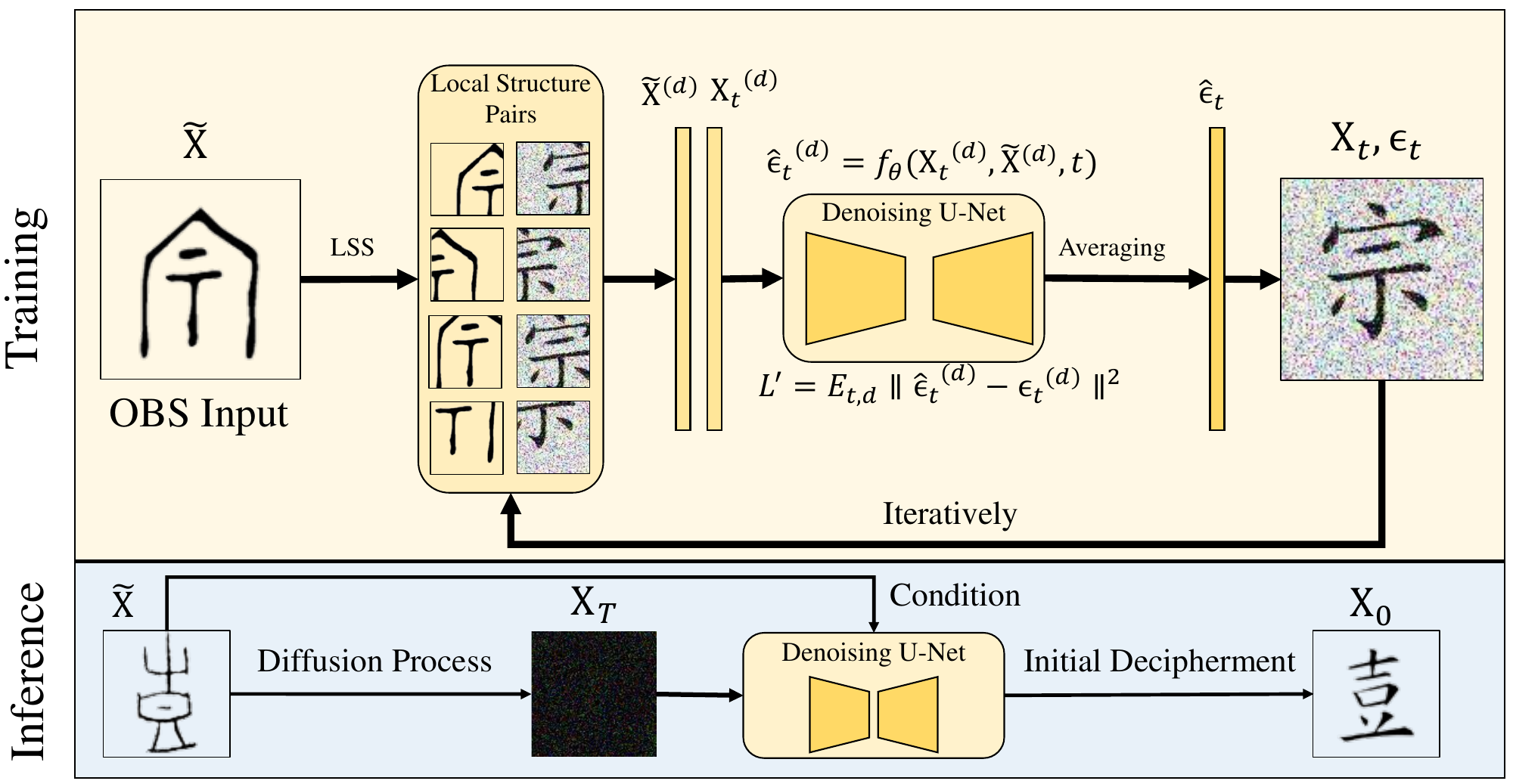}
\caption{The overview pipeline of initial decipherment of \method{}.}
\label{fig:initial-decipherment}
\end{figure*}

Figure~\ref{fig:global-local} has demonstrated that despite the considerable structural evolution from OBS to modern Chinese characters, certain local structures have been preserved. As shown in Figure~\ref{fig:initial-decipherment}, to enable the diffusion model to learn these localized radical features, the LSS module employs a sliding window approach to segment the target modern Chinese character images $X_0\in R^{H\times W\times 3}$ and corresponding OBS images $\tilde{X}\in R^{H\times W\times 3}$ into $D$ patches of size $\tilde{p}\times \tilde{p}$, denoted as $\tilde{X}^{(d)}$ and $X_t^{(d)}\in R^{\tilde{p}\times \tilde{p}\times 3}, d=1,2,...D, \tilde{p}=64$. Here, $X_{t}$ represents the modern text image with added Gaussian noise $\epsilon_{t}$ at timestep $t$. Consequently, we focus on learning the conditional reverse process as follows:
\begin{equation}
\resizebox{0.85\linewidth}{!}{$
p_\theta(X_{0:T}^{(i)}\mid\tilde{X}^{(i)})=p(X_T^{(i)})\prod_{t=1}^Tp_\theta(X_{t-1}^{(i)}\mid X_t^{(i)},\tilde{X}^{(i)})
$}
\end{equation}

By adopting this approach, the model iteratively refines each patch by learning the nuanced mappings from the localized structures of OBS to their modern counterparts. The loss function in Equation~\ref{eq:loss} can then be rewritten as follows:
\begin{equation}
    \begin{array}
        {c}\hat{\epsilon}_t^{(d)}=f_\theta(X_t^{(d)}, \tilde{X}^{(d)},t)\\ \mathcal{L'}=\mathbb{E}_{t,d}\parallel\hat{\epsilon}_t^{(d)}-\epsilon_t^{(d)}\parallel^2
    \end{array}
\end{equation}

Here, the model's goal is to minimize the difference between the estimated noise $\hat{\epsilon}_{t}^{(d)}$, and the actual noise, $\epsilon_t^{(d)}$, within each patch.

In the inference phase, our approach involves dissecting the OBS image $\tilde{X}$ into $\tilde{p} \times \tilde{p}$ patches, with $p$ set at 64, through a structured grid layout, utilizing a sliding window for systematic extraction. The grid is arranged such that each cell hosts $r \times r$ patches, with $r$ set at 16, allowing for a finer subdivision than the patch size $\tilde{p}$. Patches are extracted by navigating the grid in both horizontal and vertical directions with a step size of $r$. The initial decipherment model then progressively refines each patch by denoising and sampling.

\begin{algorithm}
\caption{LSS Algorithm}
\begin{algorithmic}[1]
\REQUIRE OBS image $\tilde{X}$, conditional diffusion model $f_\theta(X_t, \tilde{X}, t)$, dictionary of $D$ overlapping patch locations.
\STATE $X_T \sim \mathcal{N}(0, I)$
\FOR{$t = T, \ldots, 1$}
    \STATE $\Omega_t = 0$ and $M = 0$
    \FOR{$d = 1, \ldots, D$}
        \STATE $X_t^{(d)} = \text{Crop}(P_d \circ X_t)$ and $\tilde{X}^{(d)} = \text{Crop}(P_d \circ \tilde{X})$ \hfill // $P_d$ represents the mask of the $d$th patch in the image.
        \STATE $\Omega_t = \Omega_t + P_d \cdot f_\theta(X_t^{(d)}, \tilde{X}^{(d)}, t)$
        \STATE $M = M + P_d$
    \ENDFOR
    \STATE $\Omega_t = \Omega_t \oslash M$ \hfill // $\oslash$: element-wise division
    \STATE $X_{t-1} = \frac{1}{\sqrt{\alpha_t}}(X_t - \frac{1 - \alpha_t}{\sqrt{1 - \gamma_t}} \Omega_t) + \sqrt{1 - \alpha_t} \epsilon_t$ \hfill // $\epsilon_t \sim \mathcal{N}(0, I)$
\ENDFOR
\RETURN $X_0$
\end{algorithmic}
\label{alg:LSS}
\end{algorithm}

Unique to our method is the handling of overlaps between patches. Instead of waiting until the denoising is complete, we average the overlapped sections at every timestep $t$, ensuring a uniform effect across the shared areas. This continuous averaging at each timestep prevents the formation of merging artifacts that typically occur when patches are processed independently. By smoothing transitions between patches during the sampling, we avoid edge discrepancies, maintaining the visual coherence of the reconstructed image. The sampling dynamics at each step are defined by Equation~\ref{eq:sample}, which guides the process toward a seamless and artifact-free image assembly. Algorithm~\ref{alg:LSS} shows the pseudocode of LSS. Figure~\ref{fig:initial-decipherment} demonstrates the overview pipeline of initial decipherment.

\subsection{Zero-shot Refinement}

\begin{figure}[t]
    \centering
    \includegraphics[width=1\linewidth]{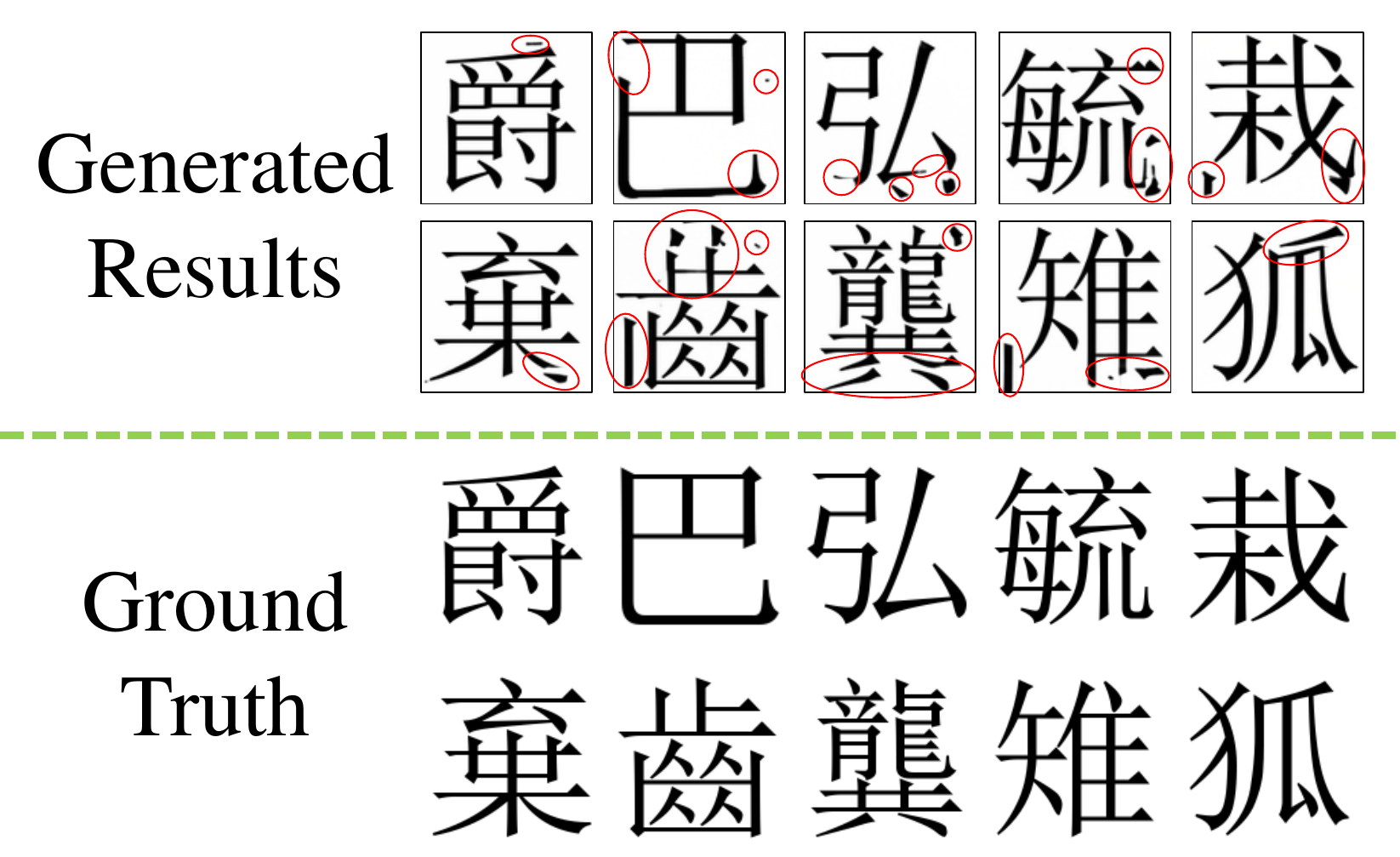}
    \caption{Modern Chinese characters generated by the initial decipherment stage, showing numerous artifacts and deformations as identified by the red circles.}
    \label{fig:initial-results}
    \vspace{-0.4cm}
\end{figure}

\begin{figure*}[t!]
    \centering
    \includegraphics[width=0.8\linewidth]{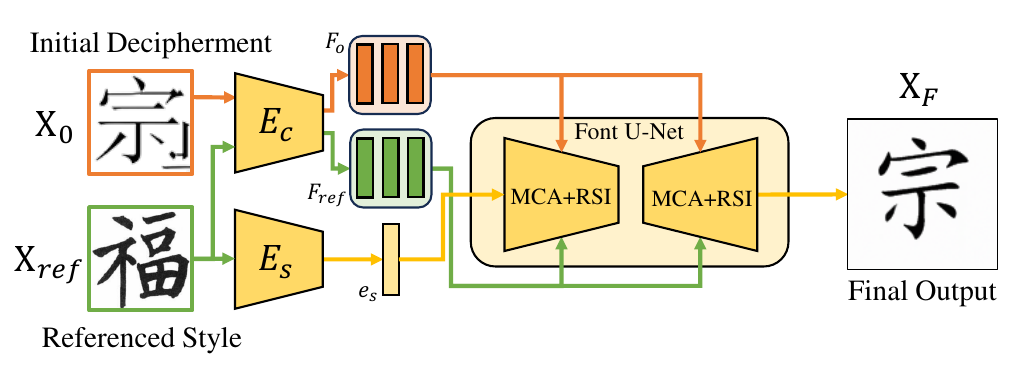}
    \caption{Overview pipeline of the zero-shot refiner.}       
    \label{fig:Zero-shot_Refiner}
    \vspace{-0.4cm}
\end{figure*}

\begin{figure}[t]
    \centering
    \includegraphics[width=0.8\linewidth]{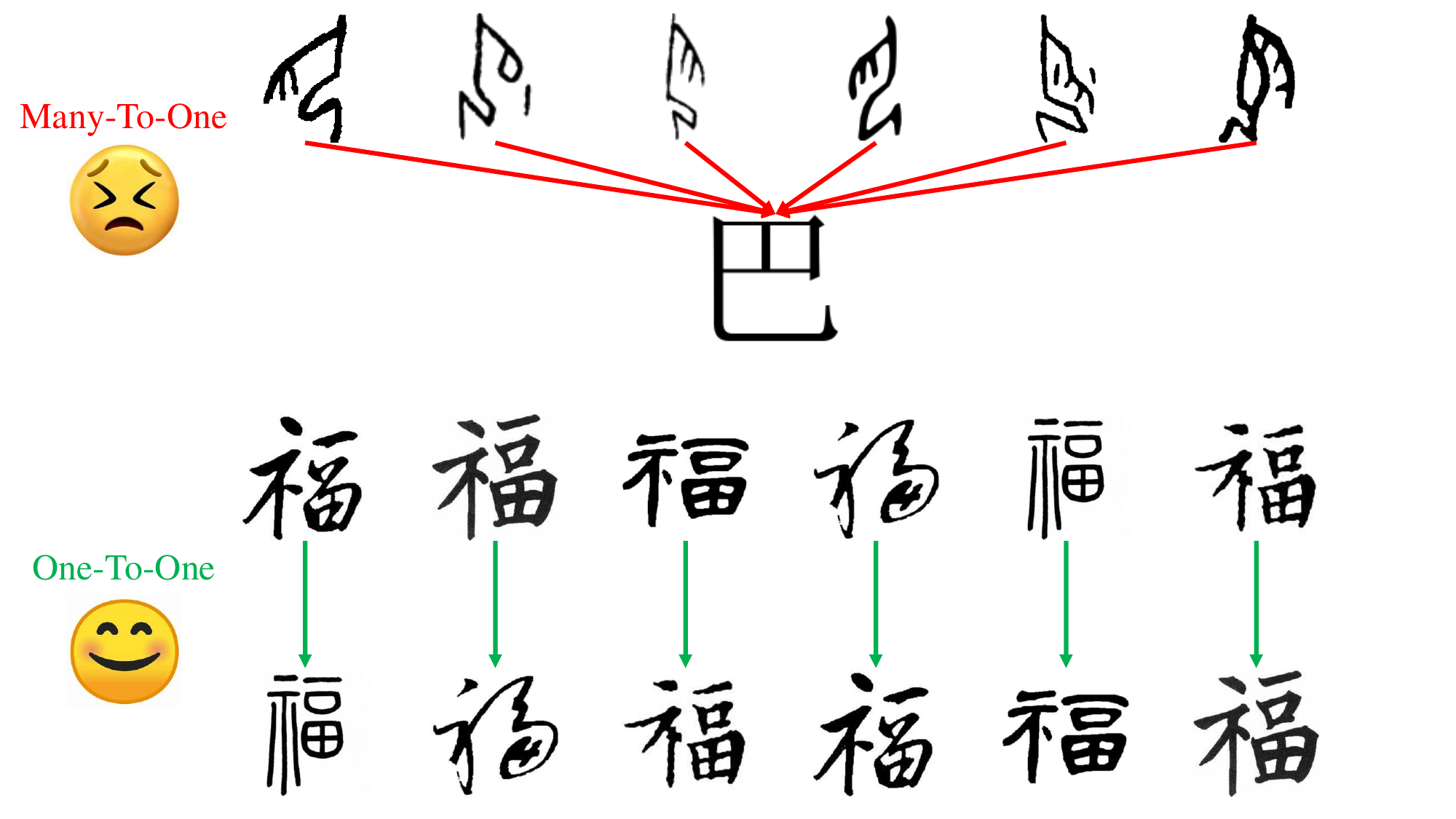}
    \caption{Comparison of many-to-one and one-to-one training paradigms. In the many-to-one approach, multiple OBC images, despite their large structural variances, are mapped to a single modern Chinese character. Conversely, the one-to-one paradigm ensures each image is individually paired.}
    \label{fig:many_one}
    \vspace{-0.4cm}
\end{figure}

Despite advancements in generating modern Chinese characters with Localized Structural Sampling, initial decipherment efforts encounter notable obstacles, such as structural deformities and artifacts, highlighted in Figure~\ref{fig:initial-results}. These issues stem from the many-to-one training approach used, where multiple OBS instances are mapped to a single modern Chinese character image (see Figure~\ref{fig:many_one}), leading to confusion and inaccuracies in capturing character evolution, and resulting in artifacts or incomplete structures due to a limited variety of modern Chinese character samples.

To overcome these challenges, we propose a zero-shot refinement strategy that involves training a model on a diverse collection of modern Chinese characters. Considering the multiple writing styles for modern Chinese characters, we aim to improve the model's understanding of their structure by employing a transformation task between different styles. We trained the module on 20 different modern Chinese character fonts to learn structural transformations between different modern Chinese character writing styles. As shown in Figure~\ref{fig:many_one}, this training process is one-to-one. This method simplifies data collection by leveraging readily available font variations, thereby enhancing the model's understanding of character structures and enabling the application of this knowledge to improve initial decipherment results without direct training on OBS-to-modern character mappings.

Our zero-shot refinement approach is grounded in a generic font style transformation framework, as depicted in Figure~\ref{fig:Zero-shot_Refiner} and based on~\cite{yang2023fontdiffuser}. The process involves a dual-encoder system to adapt the style of a source font image $X_{0}$ to a target style $X_{ref}$, preserving content integrity. The style encoder $E_s$ extracts style features $e_s$ from $X_{ref}$, while the content encoder $E_c$ processes $X_{o}$ and $X_{ref}$ to obtain multi-scale content features $F_0 = \{f_o^1, f_o^2, f_o^3\}$ and $F_{ref} = \{f_{ref}^1, f_{ref}^2, f_{ref}^3\}$, refined by a specialized UNet with Multi-scale Content Aggregation (MCA) and Reference-Structure Interaction (RSI) blocks for enhanced feature integration. The model employs cross-attention mechanisms to align features and address structural differences, formalized as:
\begin{equation}
    \begin{array}{c}
        S_{ref}\in\mathbb{R}^{C_{ref}^i\times H_iW_i}=\text{flatten}(f_{ref}^i)\\S_s\in\mathbb{R}^{C_s^i\times H_iW_i}=\text{flatten}(o_i)\\Q=\Phi_q(S_{ref}),~K=\Phi_k(S_s),~V=\Phi_v(S_s)
    \end{array}
\end{equation}

\noindent where $o_i$ represents the UNet feature derived from $f_o^i$ and $e_s$, and $\Phi_{q}$, $\Phi_{k}$, $\Phi_{v}$ denote linear projections. The deformation offset $\delta_\text{offset}$ is calculated as follows:
\begin{equation}
    \begin{array}{c}
        F_{\text{attn}} = \text{softmax}(\frac{QK^T}{\sqrt{d_k}})V\\\delta_{\text{offset}}=\text{FFN}(F_{\text{attn}})
    \end{array}    
\end{equation}

\noindent The output $I_f$ is the result of rendering the source image with DCN~\cite{dai2017deformable}, considering the calculated deformation offset:
\begin{equation}
    I_f = \text{DCN}(o_i,\delta_{\text{offset}})    
\end{equation}

In adapting the framework for OBS decipherment, we streamline the model by focusing on a singular font style, thereby omitting the style contrastive refinement module and its contrastive loss, simplifying the training process.  The encoders are trained using the offset loss $\mathcal{L}_{\text{offset}}$, which measures the mean magnitude of deformation offsets:
\begin{equation}
    \mathbf{\mathcal{L}}_{\text {offset}}=mean(\left \| \delta_{\text{offset}}  \right \| )    
\end{equation}

\noindent where $\delta_{\text{offset}}$ signifies the deformation offset, encapsulating structural information gleaned from the reference features, and the mean operation computes the average magnitude of these offsets.

After training, the zero-shot refinement module was directly employed to refine the results generated by the diffusion model.

\section{Experiments}

\subsection{Dataset and Evaluation Metric}

To train and evaluate the proposed \method{} model, we selected the HUST-OBS dataset~\cite{wang2024open} and EVOBC dataset~\cite{guan2024open}, which stands as one of the largest repositories of OBS, with 1,590 distinct characters depicted in 71,698 images. Recognizing the complexities involved in deciphering unknown OBS, which usually require comprehensive expert validation, we opted for already deciphered inscriptions in our testing set to streamline the evaluation process. Importantly, the categories of characters in the testing set were specifically chosen to be absent from the training set, ensuring that the model faces the genuine challenge of deciphering \textbf{unseen} and novel categories. The dataset was partitioned into training and test sets with a 9:1 ratio, providing a robust framework for assessment.

\begin{table*}[t!]
    \centering
    \resizebox{\linewidth}{!}{
    \begin{tabular}{c|c|ccccccc} \hline  
            Evaluation Tool &Rank&  Pix2Pix& Palette& DRIT++& CycleGAN& BBDM& CDE&OBSD (ours)\\ \hline  
            OBS-OCR&Top-1@Acc&  0.0\%& 0.0\%& 0.0\%& 0.0\%& 19.5\%& 31.0\%&\textbf{41.0\%}\\
            OBS-OCR&Top-10@Acc&  0.0\%& 0.0\%& 0.0\%& 0.0\%& 29.5\%& 47.5\%&\textbf{50.5}\%\\
            OBS-OCR&Top-20@Acc&  0.0\%& 0.0\%& 0.0\%& 0.0\%& 34.5\%& 50.0\%&\textbf{54.5}\%\\
            OBS-OCR&Top-50@Acc&  0.0\%& 0.0\%& 4.5\%& 8.5\%& 39.0\%& 52.5\%&\textbf{58.0}\%\\
            OBS-OCR&Top-100@Acc&  0.0\%& 3.0\%& 13.0\%& 19.0\%& 42.0\%& 56.0\%&\textbf{61.0}\%\\
            OBS-OCR&Top-200@Acc& 14.5\%& 8.5\%& 20.0\%& 37.5\%& 46.0\%& 59.5\%&\textbf{62.5}\%\\
            OBS-OCR&Top-500@Acc& 17.5\%& 19.5\%& 21.5\%& 60.0\%& 58.0\%& 64.0\%&\textbf{64.5}\%\\ \hline  
 PaddleOCR& Top-1@Acc& 0.0\%& 0.0\%& 0.0\%& 0.0\%& 7.0\%& 19.0\%&\textbf{30.0\%}\\ \hline
    \end{tabular}
    }
    \caption{Comparison of single-round decipherment success rate between the proposed \method{} and state-of-the-art image-to-image translation methods.}
    \label{tab:quantitative}
    \vspace{-0.4cm}
\end{table*}

While the proposed \method{} model approaches OBS decipherment from an image generation perspective, it is crucial to acknowledge that traditional image generation metrics, such as SSIM~\cite{nilsson2020understanding}, are not suitable for this distinct challenge. Instead, we adopted OCR technology as a more objective measure of decipherment success. Our custom-built OCR tool, OBS-OCR, is a simple classifier using ResNet-101 backbone specifically trained on a large dataset of 88,899 categories modern Chinese characters to evaluate the model's output. The custom-built OCR tool achieved a recognition accuracy of 99.87\% on 88,899 categories of Chinese characters, which demonstrates reliable performance to evaluate the decipherment results. Its aim is to automatically recognize the results generated by the diffusion models and compare these results with the ground truth in order to evaluate the model’s deciphering performance. By comparing the OCR-recognized characters against their ground truth labels, we simulate a quantifiable form of expert validation. To make a more reliable and objective evaluation, we also incorporated the widely-used, open-source Chinese OCR tool PaddleOCR~\footnote{\href{https://github.com/PaddlePaddle/PaddleOCR}{https://github.com/PaddlePaddle/PaddleOCR}} as an additional OCR tool to support further evaluations. This dual-OCR method provides a robust framework for assessing the model's efficacy in accurately deciphering oracle bone languages.

\subsection{Quantitative Results}

\begin{table}[t!]
    \centering
    \resizebox{\linewidth}{!}{
    \begin{tabular}{c|c|c|c|c|c|c} \hline  
             \multicolumn{2}{c|}{Number of Trial} &1&2&3&4&5\\ \hline  
             \multirow{2}{*}{Evaluation Tool}&OBS-OCR&41.0\%&56.0\%&67.5\%&75.5\%&76.5\%\\
              & PaddleOCR& 30.0\%& 40.0\%& 46.0\%& 50.5\%& 53.0\% \\  \hline
              \multicolumn{2}{c|}{Number of Trial} &6&7&8&9&10\\ \hline  
             \multirow{2}{*}{Evaluation Tool}&OBS-OCR&77.0\%&78.0\%&79.5\%&80.0\%&80.0\% \\
              & PaddleOCR& 55.5\%& 57.0\%& 57.5\%& 58.5\%& 58.5\% \\  \hline
    \end{tabular}
    }
        \caption{Top-1 accuracy of the multi-round decipherment success rate of the proposed \method{}.}
    \label{tab:Fault_tolerance}
    \vspace{-0.4cm}
\end{table}

In quantitatively evaluating the performance of our proposed \method{}, we employ two distinct assessment criteria: single-round decipherment and multi-round decipherment. The single-round decipherment evaluation aims to gauge the method's capability to decipher individual samples accurately, providing insight into its immediate effectiveness. On the other hand, the multi-round decipherment assessment offers a more practical appraisal of the method's performance, where multiple attempts at deciphering a single image are permitted. This approach mirrors the iterative nature of real-world decipherment tasks, allowing for a comprehensive assessment of the method's resilience and adaptability over successive trials.

Given the absence of dedicated tools for oracle bone language decipherment, we employ a comparative framework that adapts leading image-to-image translation methods to this specialized task. This set includes GAN-based approaches such as Pix2Pix~\cite{isola2017image}, CycleGAN~\cite{zhu2017unpaired}, DRIT++~\cite{lee2020drit++}, and diffusion-based methods like CDE~\cite{saharia2022image}, Palette~\cite{saharia2022palette}, BBDM~\cite{li2023bbdm}. This setting not only mirrors the core mechanism of our \method{} method but also allows for a comprehensive evaluation against the backdrop of the latest advancements in image translation. Each method was carefully adapted to the OBS context, ensuring consistent training and testing conditions for a fair evaluation.

In the single-round decipherment evaluation, as shown in Table~\ref{tab:quantitative}, our \method{} demonstrates a significant advantage over the adapted image-to-image translation methods in deciphering oracle bone language. Notably, the top-1 accuracy for OBS-OCR and PaddleOCR achieved by \method{} stand at 41.0\% and 30.0\%, respectively, surpassing the performance of other methods. As the rank increases, there is a clear trend of improving accuracy, at Top-500 accuracy, \method{} reaches a 64.5\% OBS-OCR recognition accuracy. It is noteworthy that all GAN-based approaches, such as Pix2Pix, Palette, DRIT++, and CycleGAN, exhibit minimal effectiveness in this context, with top-1 accuracies at 0\%. This could be attributed to the GANs' inherent challenge in capturing the complex and nuanced mappings required for accurately deciphering the oracle bone language into modern Chinese. Surprisingly, the adapted diffusion models, despite their general-purpose nature, have shown commendable performance, underscoring the viability of leveraging image generation techniques in addressing the challenges traditional NLP algorithms encounter in decipherment tasks. This aligns with our methodological premise, validating the novel approach of integrating image-based generative models into the domain of linguistic decipherment.

In addition, Table~\ref{tab:Fault_tolerance} presents the multi-round decipherment results, where a progressive increase in decipherment success rates can be witnessed across multiple trials. The OBS-OCR metric starts at a success rate of 41.0\%, and levels out at 80.0\% by the 10th trial, showcasing the cumulative benefit of iterative testing. Similarly, the PaddleOCR metric exhibits a consistent upward trend, commencing at 30.0\% and culminating at 58.5\% in the final trial. These results validate the incremental improvements achievable through successive attempts.

\subsection{Ablation Study}

To further examine the impact of individual components in our proposed method, we conducted an ablation study focusing on the LSS module and zero-shot refinement. The results, presented in Table~\ref{tab:ablation}, highlight the limitations of employing only the basic conditional diffusion model for OBS decipherment, which resulted in notably low accuracy rates. Specifically, training the diffusion model without any enhancements led to outputs that were essentially nonsensical, characterized by random and uninterpretable stroke combinations (see Figure~\ref{fig:failure}). The introduction of the LSS module marked a significant improvement, enabling the generation of decipherment outcomes with a Top-1 recognition rate of 37.5\% for OBS-OCR and 24\% for PaddleOCR. The addition of the zero-shot refinement module, in conjunction with the LSS, further increased the Top-1 accuracy for both OBS-OCR and PaddleOCR by an additional 3.5\% and 6\%, respectively.

\begin{table}[t]
\centering
\resizebox{\linewidth}{!}{
\begin{tabular}{c|c|c|c|c}
\hline
Metric    & \multicolumn{1}{c|}{Rank} & Diffusion & +LSS & +Refinement \\ \hline
OBS-OCR   & Top-1                 & 0.5\%       & 37.5\%            & 41.0\%                                    \\
OBS-OCR   & Top-10                & 2.5\%       & 49.0\%              & 50.5\%                                  \\
OBS-OCR   & Top-20                & 4.5\%       & 52.0\%              & 54.5\%                                  \\
OBS-OCR   & Top-50                & 6.5\%       & 55.0\%              & 58.0\%                                    \\
OBS-OCR   & Top-100               & 9.0\%       & 58.0\%              & 61.0\%                                    \\
OBS-OCR   & Top-200               & 10.5\%       & 60.5\%            & 62.5\%                                  \\
OBS-OCR   & Top-500               & 16.5\%       & 64.0\%              & 64.5\%                                  \\ \hline
PaddleOCR & Top-1                 & 0.0\%       & 24.0\%              & 30.0\%                                    \\ \hline
\end{tabular}
}
\caption{Ablation Study of \method{}.}
\label{tab:ablation}
\vspace{-0.4cm}
\end{table}

\subsection{Qualitative Results}

\begin{figure}[ht!]
\centering
\includegraphics[width=\linewidth]{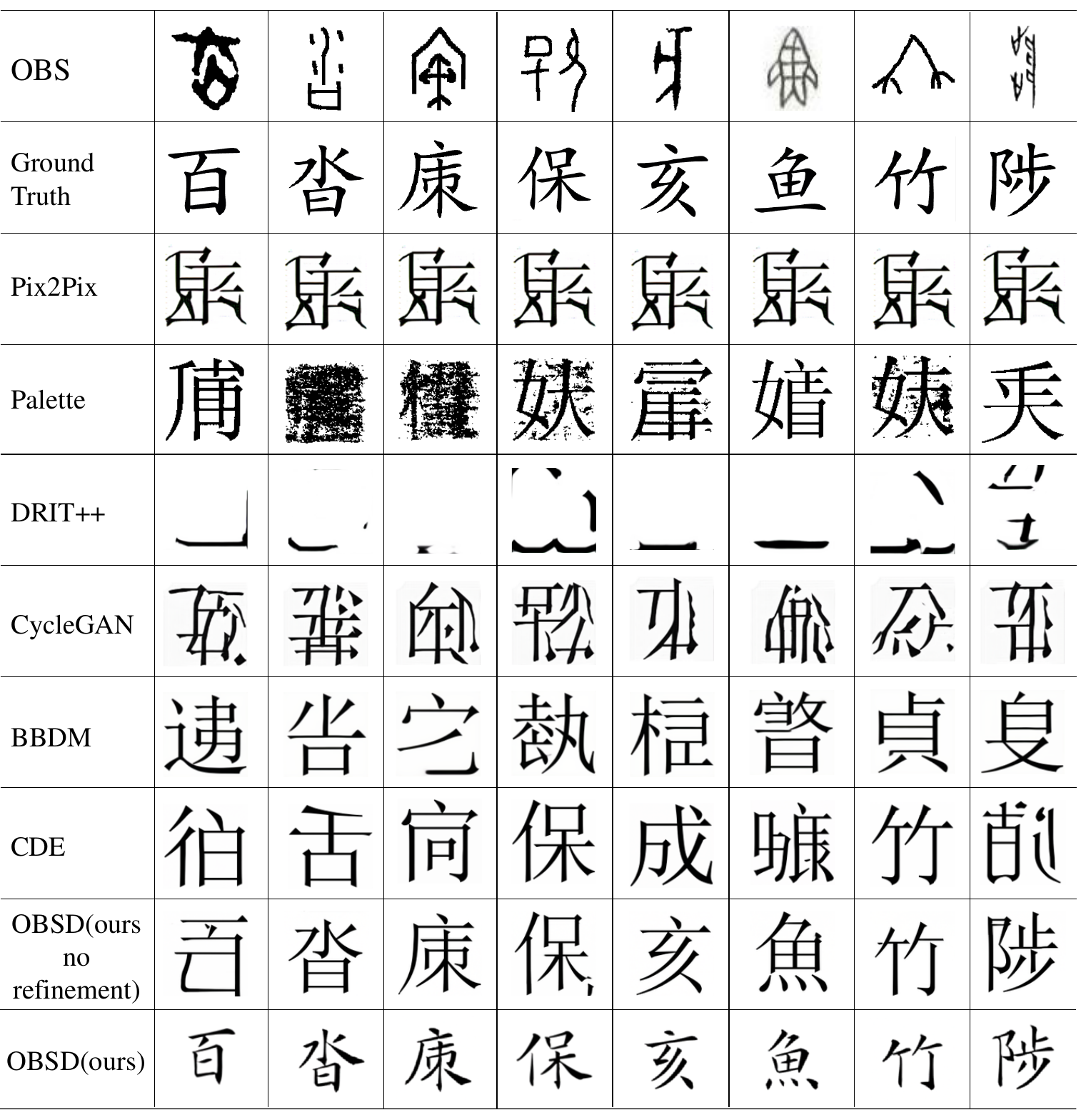}
\caption{Comparison of qualitative results between the proposed OBSD and other state-of-the-art image-to-image translation frameworks, including Pix2PIx~\cite{isola2017image}, Palette~\cite{saharia2022palette}, DRIT++~\cite{lee2020drit++}, CycleGAN~\cite{zhu2017unpaired}, BBDM~\cite{li2023bbdm}, and CDE~\cite{saharia2022image}.}
\label{fig:qualitative}
\end{figure}

Figure~\ref{fig:qualitative} showcases the qualitative results of various image-to-image translation models, with our method, \method{}, standing out by producing the most accurate reconstructions of modern Chinese characters from OBS inputs. Pix2Pix~\cite{isola2017image}, for example, generates outputs that are highly uniform across different inputs, demonstrating a lack of differentiation in character decipherment. On the other hand, DRIT++~\cite{lee2020drit++} struggles to produce complete characters, often resulting in fragmented and unrecognizable forms. In stark contrast, \method{} demonstrates a robust capability to discern and reconstruct the intricate details of each OBS, leading to coherent and precise character forms that closely align with the ground truth. These results not only highlight the efficacy of \method{} but also its potential as a tool for experts in the field of oracle bone language decipherment.

To demonstrate the performance of \method{} on authentic, undeciphered OBS, we present an extensive evaluation in the appendix, depicted in Figure~\ref{fig:D_1},~\ref{fig:D_2},~\ref{fig:D_3} and~\ref{fig:D_4}. This evaluation showcases a range of decipherment outcomes, from partial reconstructions that shed light on the structural elements of OBS characters, such as radicals and strokes, to complete character forms that exhibit a high resemblance to modern Chinese script. While the bulk of these results provide structural clues, the fully reconstructed characters hold particular promise, indicating the potential of \method{} to contribute meaningfully to the field of oracle bone language decipherment.

\subsection{Discussion}

\textbf{Experiment Results:} We compared the proposed OBSD with other generic image generation models for the OBS deciphering task. As shown in Figure~\ref{fig:qualitative}, most generic image generation models fail to produce structurally complete Chinese characters. This is because these methods, based on conditional generation, attempt to directly map the input OBS image to modern characters, neglecting the structural and writing conventions of the characters. In contrast, the proposed OBSD addresses these issues by incorporating local radical structure information into the training process, resulting in more accurate outputs.

\noindent 
\textbf{Analysis of Proposed Modules:} According to the experimental results, we found that the proposed LSS module effectively directs the diffusion model's focus towards the local structures of both OBS and modern Chinese characters. This results in clearer character strokes and more reasonable character structures. Additionally, the Zero-shot Refinement module refines the initial decipherment results by learning the structural characteristics of modern Chinese characters, ensuring a more precise and coherent structure.

\noindent 
\textbf{Generalizability to Other Languages:} The proposed method was initially designed for ideographic or pictographic languages, such as Chinese characters or Mayan script, where a single character represents a word or morpheme. This design enables the adaptation of the method to similar languages. For alphabetic scripts, which typically have a small number of letters, decipherment is rarely an issue. The applicability of these methods to other languages presents an interesting research question, which we will explore in our future work.

\section{Conclusion}

In this work, we presented \method{}, an innovative approach leveraging conditional image generation for the decipherment of OBS. Our novel Local Structure Sampling technique addresses the inherent challenges in learning modern Chinese characters' structures from limited samples, enabling effective structural correspondence learning between OBS and modern Chinese characters. Furthermore, the integration of a zero-shot refinement module significantly enhances the decipherment accuracy, a claim substantiated by promising results on the HUST-OBS dataset and EVOBC dataset. The potential of \method{} extends beyond OBS, offering prospects for deciphering other ancient scripts, such as hieroglyphs and Maya glyphs. Looking ahead, we aim to collaborate with epigraphy experts to further validate and refine the \method{}, aspiring to advance AI's role in the decipherment of ancient languages.

\section{Limitations}

In this study, we employed OCR technology, including a custom-built tool and the off-the-shelf package PaddleOCR, to evaluate the success of our \method{} in deciphering oracle bone language. While this approach offers a novel and objective metric, it is important to recognize its inherent limitations. However, these methods cannot be directly applied to evaluate truly undeciphered OBS, where the absence of ground truth necessitates expert validation.

Evaluating the decipherment results of entirely unknown OBS characters presents a unique challenge that goes beyond the capabilities of OCR technology. This task involves interpreting historical, cultural, and linguistic contexts that are deeply embedded within the languages. Therefore, the ultimate validation of our model's decipherment for such inscriptions requires the involvement of scholars and experts in oracle bone studies. We acknowledge the importance of this expert validation and are exploring collaborations with specialists in the field to assess the relevance and accuracy of our model's outputs for genuinely undeciphered texts.

\section{Acknowledgments}
This work was supported by the National Natural Science Foundation of China (No. 61936003, No.62225603, No.62206103, No.62441604).

\bibliography{ref}

\appendix
\newpage
\section{Appendix}
\label{sec:appendix}

\subsection{Implementation Details}

The proposed \method{} was trained using the Adam optimizer with a weight decay of $10^{-4}$, $\beta_{1}=0.9$, and $\beta_{2}=0.999$. During training, the learning rate was set to $2e^{-5}$, and the batch size was 8. Each batch contained 8 patches of size $64 \times 64$, and the model was trained on an Nvidia RTX A6000 model for 300 epochs. The entire training process spanned over 2 weeks.

\subsection{Decipherment Results on Genuine Unknown OBS}

\begin{figure}[ht]
\centering
\includegraphics[width=\linewidth]{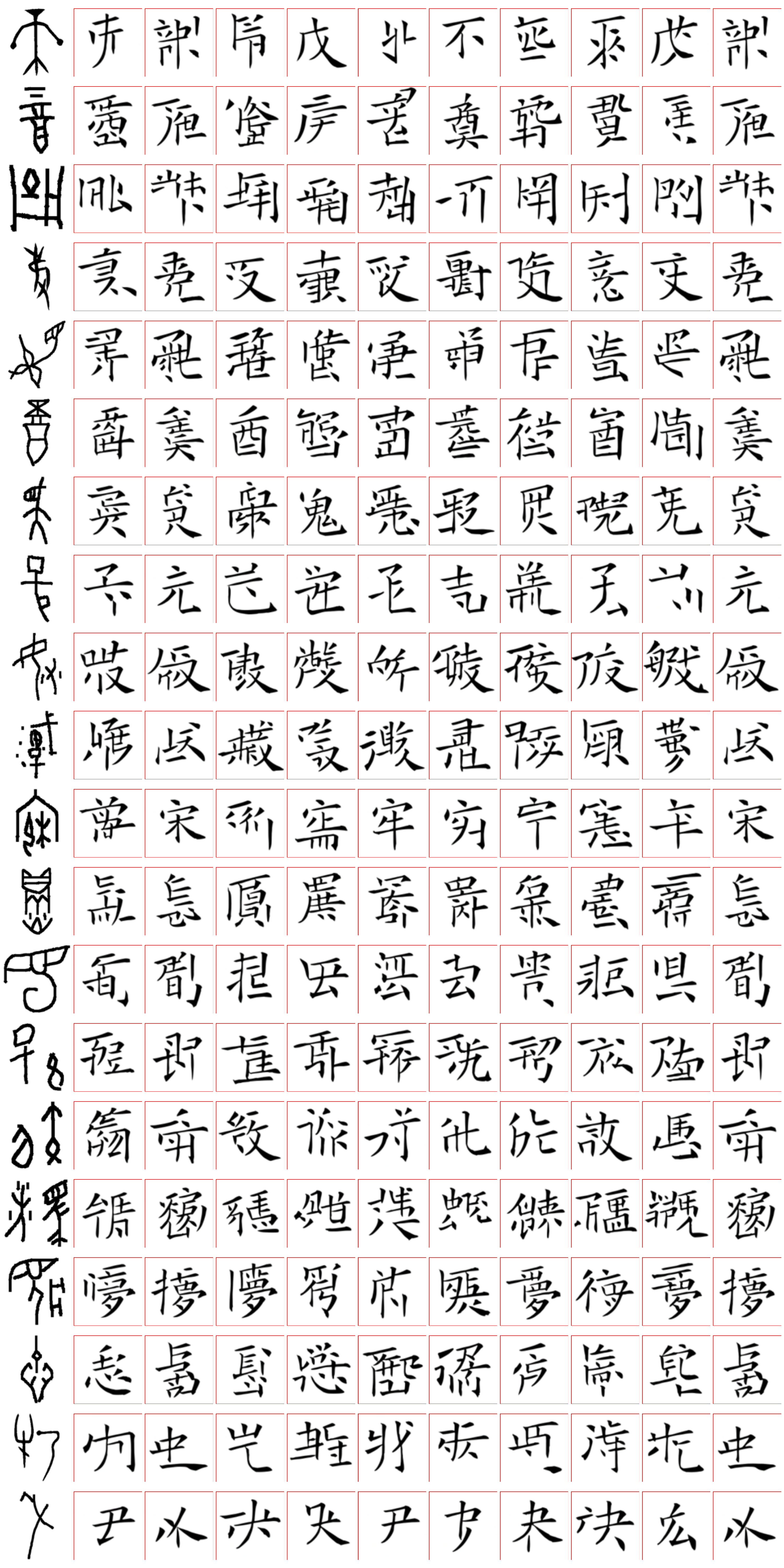}
\caption{Deciphered results for genuine undeciphered OBS.}
\label{fig:D_1}
\end{figure}

\begin{figure}[ht]
\centering
\includegraphics[width=\linewidth]{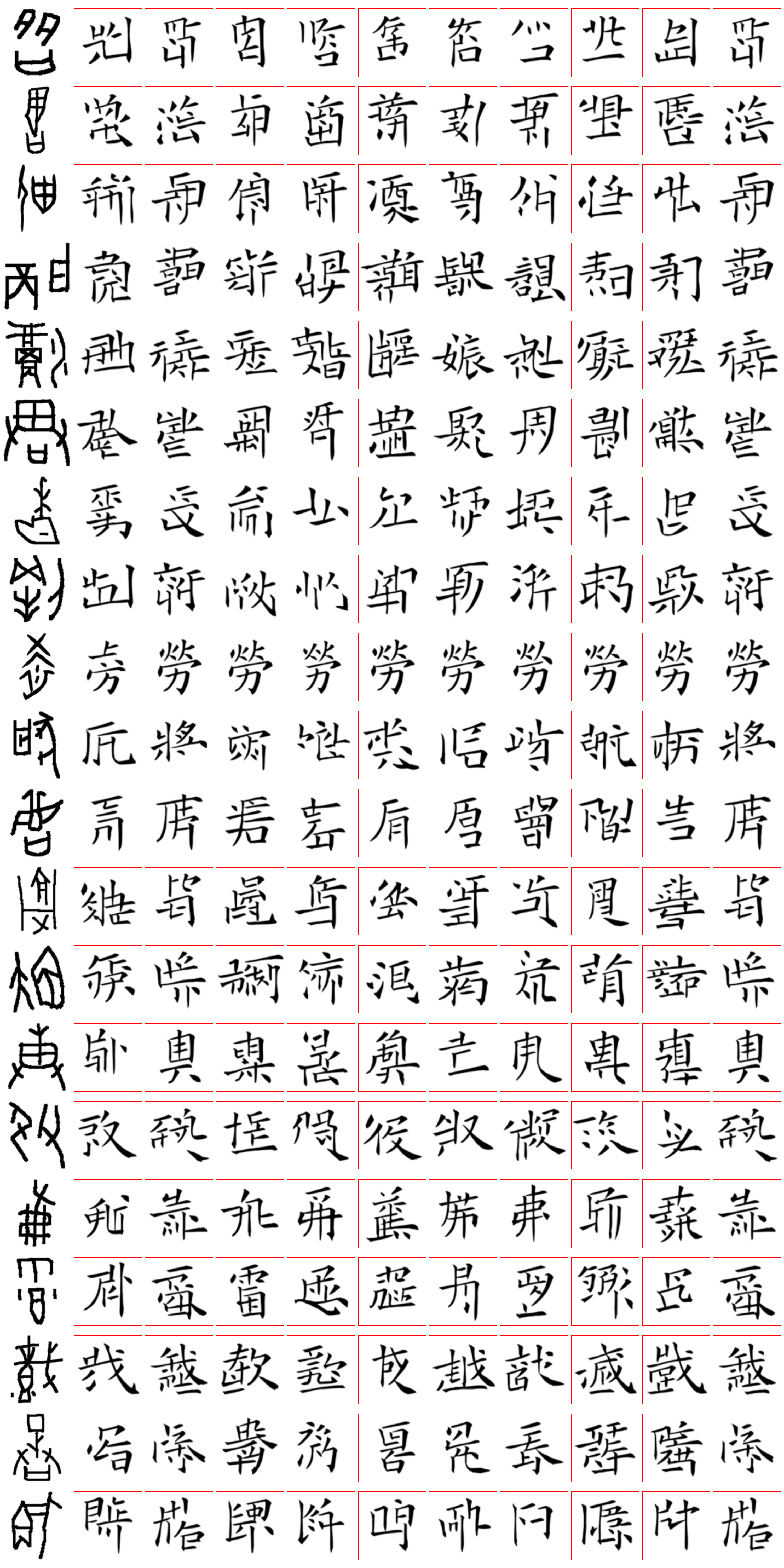}
\caption{Deciphered results for genuine undeciphered OBS.}
\label{fig:D_2}
\end{figure}

\begin{figure}[ht]
\centering
\includegraphics[width=\linewidth]{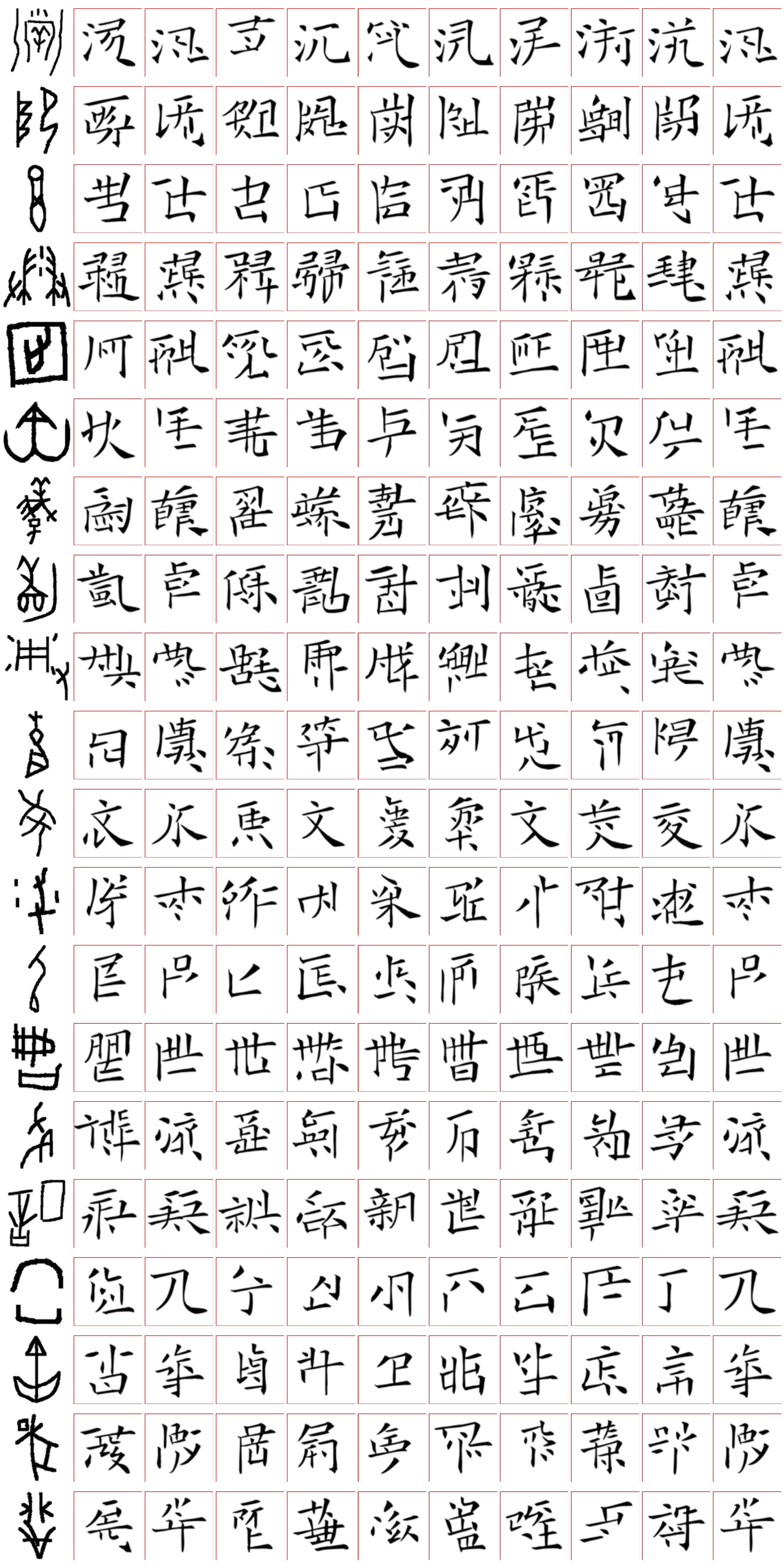}
\caption{Deciphered results for genuine undeciphered OBS.}
\label{fig:D_3}
\end{figure}

\begin{figure}[ht]
\centering
\includegraphics[width=\linewidth]{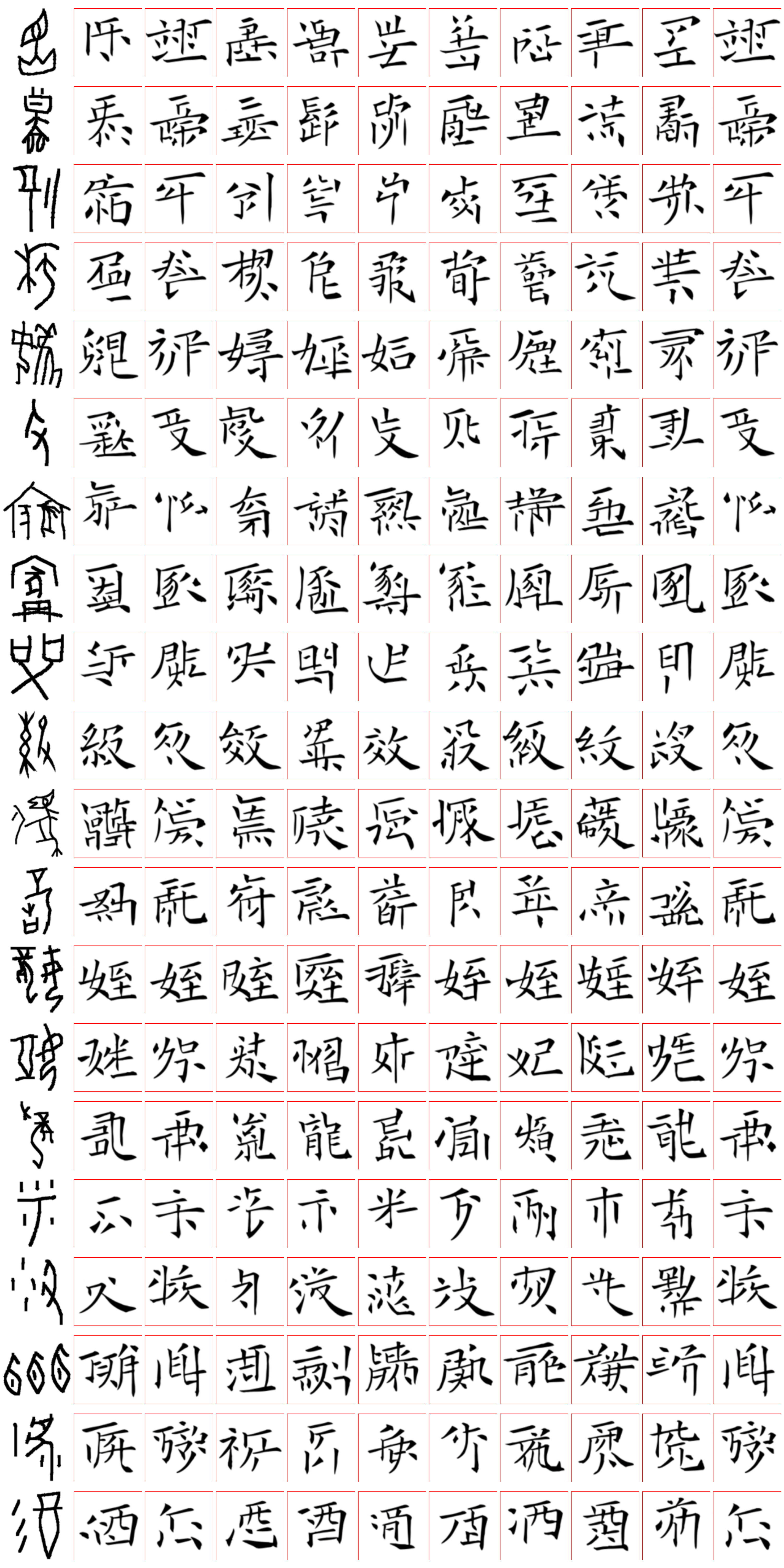}
\caption{Deciphered results for genuine undeciphered OBS.}
\label{fig:D_4}
\end{figure}

Figure~\ref{fig:D_1},~\ref{fig:D_2},~\ref{fig:D_3} and~\ref{fig:D_4} showcase the \method{} model's decipherement outputs for previously undeciphered characters from the HUST-OBS~\cite{wang2024open} dataset and EVOBC dataset~\cite{guan2024open}. For each character, we present a set of 10 potential interpretations, generated using distinct random seeds to ensure diversity in the results. In our commitment to supporting ongoing research in this field, we plan to make the code, the pre-trained models, and a comprehensive collection encompassing all decipherment outcomes publicly available. We hope this contribution will assist scholars and researchers in advancing the study of ancient languages.

\newpage

\newpage

\newpage

\newpage

\end{document}